# Probe-based Rapid Hybrid Hyperspectral and Tissue Surface Imaging Aided by Fully Convolutional Networks

Jianyu Lin [1,2], Neil T. Clancy [1,3], Xueqing Sun [1,3], Ji Qi [1,3], Mirek Janatka [4,5], Danail Stoyanov [4,5], Daniel S. Elson[1,3]

[1] Hamlyn Centre for Robotic Surgery, Imperial College London, UK
[2] Department of Computing, Imperial College London, UK
[3] Department of Surgery and Cancer, Imperial College London, UK
[4] Centre for Medical Image Computing, University College London, UK
[5] Department of Computer Science, University College London, UK

**Abstract.** Tissue surface shape and reflectance spectra provide rich intra-operative information useful in surgical guidance. We propose a hybrid system which displays an endoscopic image with a fast joint inspection of tissue surface shape using structured light (SL) and hyperspectral imaging (HSI). For SL a miniature fibre probe is used to project a coloured spot pattern onto the tissue surface. In HSI mode standard endoscopic illumination is used, with the fibre probe collecting reflected light and encoding the spatial information into a linear format that can be imaged onto the slit of a spectrograph. Correspondence between the arrangement of fibres at the distal and proximal ends of the bundle was found using spectral encoding. Then during pattern decoding, a fully convolutional network (FCN) was used for spot detection, followed by a matching propagation algorithm for spot identification. This method enabled fast reconstruction (12 frames per second) using a GPU. The hyperspectral image was combined with the white light image and the reconstructed surface, showing the spectral information of different areas. Validation of this system using phantom and *ex vivo* experiments has been demonstrated.

## 1 Introduction

Tissue surface shape measurement is a tool for both surgical navigation and pathology detection. For example, morphological appearance could assist in colonic polyp detection [1]. In addition intra-operative tissue surface shape can be combined with pre-operative imaging modalities like CT or MRI, aiding surgical navigation [2]. SL is an active stereo technique used for surface reconstruction, provides similar reconstruction accuracy to passive stereo, and outperforms methods like shape-from-shading and time-of-flight. Due to its non-reliance on object surface texture, SL has shown potential in textureless tissue surface reconstruction in surgical environments [2]. A typical SL system consists of a projector and camera. A pattern decoding step finds the correspondences between the camera and/or projector image planes, and enables surface reconstruction by triangulation for a specific camera-projector position.

Spectral imaging techniques including HSI and multispectral (MSI) implementations, measure the reflectance spectra at particular locations in an image, and have useful clinical applications in discriminating tissues which are indistinguishable under white light [3]. MSI has been used to monitor perfusion and tissue oxygen saturation intra-operatively in the bowel [4], during transplant surgery, and vascular procedures [5]. The high resolution of HSI, compared to multispectral imaging (MSI), can also enable more quantitative analysis allowing extraction of structural information or identification of disease markers in tissue using machine learning algorithms [6]. Previously combination of spectral and 3D information using MSI systems has been attempted [7, 8], but acquisition time has limited the spectral resolution to tens of wavelengths or less. To the best of our knowledge, this is the first time HSI (hundreds of wavelengths) has been combined with 3D reconstruction techniques based on SL and we believe that this solution to the compromise of spectral versus spatial resolution can allow highly accurate rapid distinction between different pathologies.

Computational image analysis involves the development of a robust near real-time algorithm to decode the projected pattern for 3D tissue surface reconstruction, based on deep learning and a pattern-specific feature matching method. Recently, advances in hardware have made solving large-scale problems using Convolutional Neural Networks (CNN) possible in a reasonable amount of time. Long et al. proposed to use Fully Convolutional Networks (FCN) for pixelwise semantic segmentation, exceeding the state-of-the-art segmentation [9]. FCN has the benefits of non-dependency on the manually extracted features, combining both global and local information, and fast execution time. In the field of microscopic imaging, FCN has been used to detect and segment cells successfully [10]. So in this work FCN is applied for pattern decoding to detect projected spots. We also show that even with a limited amount of training images (n=17) by data augmentation FCN still performs robustly and accurately.

## 2 Materials and Methods

### 2.1 Hybrid Structured Light and Hyperspectral Imager (SLHSI)

The system is built around a custom optical fibre assembly (Fibertech Optica, Inc., Canada) similar to that in our previous work [11, 12]. It is a 2.5 m long incoherent bundle of 171 50 μm core fibres arranged in a linear array at one end and a circular bundle at the other (Fig. 1). A 20 mm working distance GRIN lens (GRINtech GmbH, Germany) is attached distally. The probe's outer diameter is 2.8 mm.

**SL.** A 4 W supercontinuum laser, dispersed by a prism, is coupled into the probe's linear array (Fig. 1). The GRIN lens projects an image of the bundle's end face which, due to the incoherent fibres, is a mixture of multi-coloured spots. The pattern is reflected by the object, collected by a laparoscope (Karl Storz GmbH, Germany) and imaged onto a CCD (Prosilica GX1050C; Allied Vision Technologies, Inc., USA).

**HSI.** With the laser switched off a white light source (Xenon 300; Karl Storz GmbH, Germany) is activated. Reflected light is collected by the probe and directed, via a 45° mirror on a motorised flipper mount (MFF101; Thorlabs Ltd., UK), towards a 250/50 mm focal length lens combination to form a demagnified image of the fibre array on the slit of a HSI camera (Nano-Hyperspec; Headwall Photonics, Inc., USA) (Fig. 1).

**Correspondences between SL and HSI.** The incoherent bundling of fibres in the probe meant that the mapping between an individual fibre's HSI spectral line and its position in the object plane had to be determined. This was done by projecting the SL pattern onto a white screen and imaging it with a separate hyperspectral camera [4] to determine each spot's mean wavelength. These could then be linked to the corresponding HSI sensor locations by sorting their wavelengths from shortest to longest.

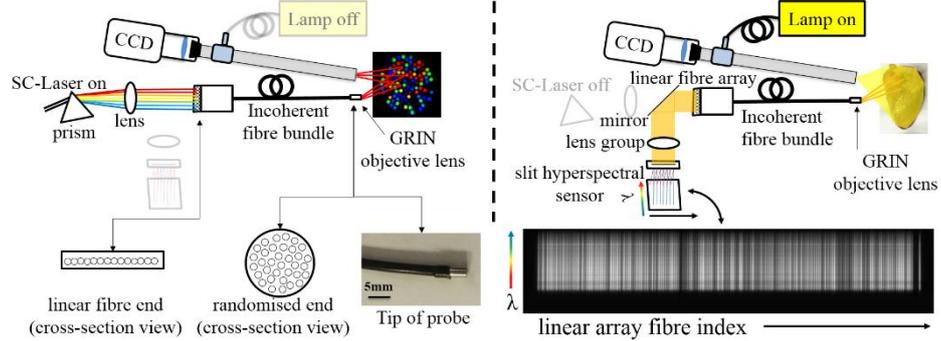

**Fig. 1.** Left: setup in SL mode, and the physical appearance of the miniaturised probe (lower left). Right: setup in HSI mode and the captured spectrograph image (lower right)

**Spectral processing.** For each fibre in the HSI image its corresponding pixel columns were averaged to reduce signal noise. The linear wavelength-pixel row relationship was calibrated by illuminating the probe with laser light of known wavelengths and recording the positions of the intensity maxima. Each fibre's spectrum was divided by the corresponding signal from a white reference target (Spectralon; Labsphere, Inc., USA) to correct for wavelength-dependent transmission characteristics of the system. Further processing to generate absorbance spectra and determine relative haemoglobin concentration in tissue was performed as described previously [4].

### 2.2 3D tissue surface reconstruction

**SL system calibration.** During calibration the projector can be regarded as an inverse camera. The calibration method has been described previously [13]. Besides, a virtual projector image (reference image) is generated from an SL pattern on a white plane using the probe intrinsic parameter and the homography [14]. It functions like the image from the second camera in passive stereo, facilitating pattern decoding.

**Pattern decoding – spot detection.** In this system projected spots with different colour are considered as features. In surgical environments spot detection is the key challenging step for accurate pattern decoding. An algorithm based on FCN has been employed to detect the spots in endoscopic images robustly in near real-time despite confounding factors such as blood, smoke or non-uniform illumination.

Our FCN model (Fig. 2 (a)) consists of two parts: the contractive and the expansive phases. The former halves image size and doubles the feature dimension at each step. At the end of this phase, each neuron has an effective receptive field of 16×16 pixels, and the feature dimension of its output volume is 512. The expansive phase also contains four steps, where each step upsamples image, fuses the upsampled output and the convoluted output from the counterpart in the contractive phase, and then convolutes to decrease the feature dimension. This expansive phase thus combines the coarse "what" and fine "where" information, resulting in a pixel-wise prediction of the whole image. The output of this model is a 2-channel image with the same size as the input image. Each channel indicates the "probabilities" of being foreground (spots) or background. The loss function is evaluated using the softmax function.

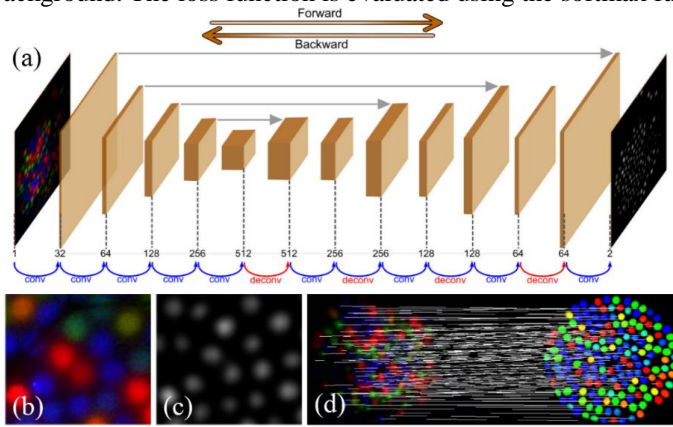

**Fig. 2.** (a) The deployed FCN model; (b) Cropped SL image; (c) Density map of the cropped SL image; (d) Feature matching between the captured (left) and the reference SL image (right)

For training 17 captured images have been used and image argumentation including resizing, flipping, and rotation, was applied to increase the training set. Input image size was halved to increase speed. During training the manual segmentations of the SL images were used as the desired label images. The training included a coarse training and a fine tuning, each with momentum 0.9 and weight decay 0.0005. The former had a learning rate of 0.0000005 until 80000 iterations, while the latter had 0.0000001 until 150000 iterations. In prediction the subtraction between the foreground and background channels was used as the "density map" of spot detection (Fig. 2 (c)). Both training and prediction were applied using Caffe [15]. In prediction the spot centres were detected at the local maxima locations of the density map.

**Pattern decoding- spot matching.** Like feature matching in passive stereo, detected spots should be matched to counterparts in the reference image for 3D reconstruction. On the reference image the spots were manually segmented offline.

The spot matching algorithm is based on colour and neighbourhood information detected using Delaunay triangulation. Neighbourhood areas of spots were defined according to distances between neighbours. Then a customised feature descriptor was defined for each spot. Taking the spot as the area centre, its feature descriptor was a 32×3 matrix, each row of which represented the colour in one direction on the area boundary. Only one feature descriptor was generated per spot in the captured SL image, while 32 descriptors starting from different directions were generated per spot in the reference image, taking rotation into consideration.

Epipolar lines were used to constrain the matching search space. The smallest distance between a spot descriptor on the captured image and all 32 descriptors of a spot on the reference image was used to describe the distance between them. The match with closest distance smaller than a threshold was chosen. This was followed by a pruning procedure based on neighbourhood information. An iterative method was then applied: matching was propagated to other neighbouring unmatched spots, followed by pruning, until the number of matches stops changing.

**Combination of SL and HSI.** With calibration and spot matching the 3D tissue surface could be triangulated. Hyperspectral data from different fibres, corresponding to individual spots, could be projected onto the reconstructed surfaces, providing a hybrid view of both the spectral and shape information relating to the target tissue. The computation time for reconstruction from single SL image was ~80 ms on a PC (OS: Ubuntu 14.04; processor: i7-3770; graphics card: NVIDIA GTX TITAN X).

## 3    Experiments and Results

Phantom and *ex vivo* tissue experiments have been carried out to validate the SL reconstruction and demonstrate hybrid hyperspectral and surface shape imaging.

**SL Reconstruction.** In SL mode the angle between the probe and the camera is ~10-15°, baseline ~3 cm, working distance 5-9 cm, with the endoscope optical axis roughly perpendicular to the tissue surface. Previous work has shown that the projected spot patterns are robust to changes in background albedo, due to their narrow bandwidth [11]. It was also found that, given an accurate calibration and perfect feature matching, reconstruction error can reach 0.7 mm at a working distance of ~10 cm [13]. Thus, the reconstruction accuracy mainly depends on the pattern decoding results. Here we provide the validation of feature matching from a silicone heart phantom, and *ex vivo* experiments on ovine heart and liver. Each group of validation data contains 10 images chosen randomly from recorded videos. Automatic feature matching results were compared with those from manual annotation. The true positive, annotated matches, together with the matching sensitivity and precision are listed in Table 1.

Table 1. Validation of feature matching. Manual annotation is used as the ground truth.

| Object | Annotated matches | True positive | Sensitivity | Precision |
|---|---|---|---|---|
| Silicone heart phantom | 170 ±1 | 154 ±17 | 0.907 ±0.101 | 0.997 ±0.004 |
| Ovine heart | 134 ±8 | 113 ±10 | 0.844 ±0.038 | 0.997 ±0.006 |
| Ovine liver | 128 ±11 | 110 ±13 | 0.861 ±0.052 | 0.994 ±0.005 |

Table 1 shows that the pattern decoding algorithm functions robustly with both the phantom and even some challenging *ex vivo* data, with feature matching sensitivity higher than 0.8, and precision 0.99. Compared with previous work [13], this shows a much higher sensitivity for feature matching due to the high spot detection accuracy achievable with FCN. Meanwhile, the near real-time performance of FCN guarantees its real-world practicality. The specified feature descriptor which accounts for pattern rotation, not only functions robustly but also simplifies the matching procedure.

**HSI validation.** Reflectance spectra from spot regions on a Macbeth colour chart, indicated by the SL pattern in Fig. 3(a), are plotted in Fig. 3(b) alongside those measured by a spectrometer (USB4000HR; Ocean Optics, Inc., USA). In each panel the SLHSI system's mapping procedure returns the correct reflectance spectrum for each location, with a mean spectral error of 10% between the SLHSI and the gold standard.

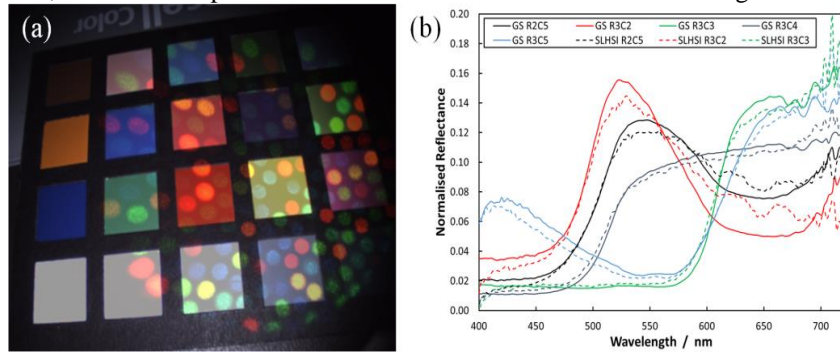

**Fig. 3.** (a) Macbeth colour chart illuminated with SL. (b) Reflectance spectra from selected colour panels of the chart, with *R* and *C* indicating the row and column, respectively. Data from the hybrid system (SLHSI) are plotted alongside the high resolution spectrometer results (GS).

To demonstrate combined shape and spectral information acquisition a cylindrical three-coloured target was imaged (Fig. 4). Mean reflectance curves for each region show peaks in the red, green and blue, and convolution with a colour CCD's spectral sensitivity allows RGB data recovery, shown in the surface mesh overlay (Fig. 4 (c)).

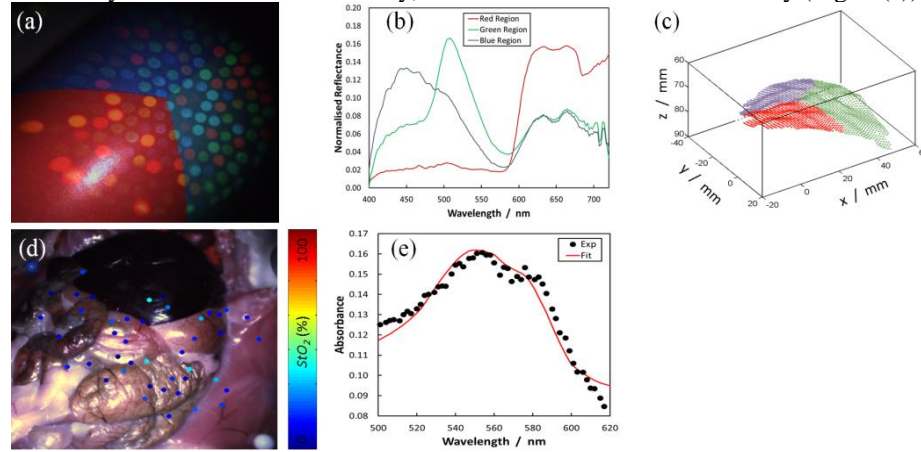

**Fig. 4.** (a) Three-coloured cylindrical target with SL. (b) Mean reflectance spectra for each of the target's regions. (c) 3D reconstruction coloured by RGB values generated from spectra in (b). (d) Murine abdomen with $StO_2$ overlay. (e) Tissue absorbance spectrum and model fit.

Reflectance spectra from a murine abdomen, imaged *post mortem*, were used to calculate the absorbance spectrum for each spot location and extract tissue oxygen saturation (StO2) using a method described previously [4]. StO2 at various spot locations is shown in Fig 4 (d), along with an absorbance spectrum from one region (Fig. 4 (e)). Spectra that did not match the model well ($r2<0.8$) were rejected. Mean StO2 in the abdomen is low ($14\pm10\%$), as expected for an *ex vivo* experiment.

According to the colour chart validation, this system can measure sample reflectance spectra accurately. This conclusion should apply to other objects including *in vivo* tissue. The main differences to be expected *in vivo* are higher StO2 values and breathing/peristalsis-related movement. The difference in StO2 will only affect the measured spectral shape. Tissue motion will not introduce much spectral artefact as each fibre's signal is acquired in a single snapshot (100 ms).

## 4      Discussion and Conclusion

We have developed a flexible rapid hybrid system capable of 3D sensing and HSI, using a motorised flipper mirror to switch between modes. The probe's size and flexibility means that it is compatible with standard clinical endoscopic tools for assessing the gastrointestinal tract and abdomen, while its imaging capabilities can enable clinical studies based on previous work in measurement of oxygenation dynamics [4], tissue classification [3] and augmentation of the clinician's view with pre-operative data [2].

SL mode enabled surface reconstruction of up to 171 data points, using a random-coloured spot pattern; while HSI measured the reflectance spectra of the same regions in one exposure. In addition, an accurate and fast 3D reconstruction algorithm was proposed, using FCN with specific feature descriptors, resulting in near real-time measurement of tissue surface even with low quality images. Future work will focus on system reliability and practicality enhancements, including compatibility with existing clinical instruments and sterilisability. Pilot studies on freshly excised tissue from human surgical procedures are planned and will allow testing of the system against histology, while further preclinical work will enable evaluation and optimisation of performance *in vivo*.

**Acknowledgements.** This work is funded by ERC 242991 and an Imperial College Confidence in Concept award. Jianyu Lin is supported by IGHI scholarship. Neil Clancy is supported by Imperial College Junior Research Fellowship. Danail Stoyanov is funded by EPSRC (EP/N013220/1, EP/N022750/1, EP/N027078/1, NS/A000027/1) and the EU-Horizon2020 (H2020-ICT-2015-688592).

## References

1. Schwartz, J.J., Lichtenstein, G.R.: Magnification endoscopy, chromoendoscopy and other novel techniques in evaluation of patients with IBD. Techniques in Gastrointestinal Endoscopy 6, 182-188 (2004)

2. Maier-Hein, L., Mountney, P., Bartoli, A., Elhawary, H., Elson, D., Groch, A., Kolb, A., Rodrigues, M., Sorger, J., Speidel, S., Stoyanov, D.: Optical techniques for 3D surface reconstruction in computer-assisted laparoscopic surgery. Medical Image Analysis 17, 974-996 (2013)
3. Lu, G., Fei, B.: Medical hyperspectral imaging: a review. Journal of Biomedical Optics 19, 010901-010901 (2014)
4. Clancy, N.T., Arya, S., Stoyanov, D., Singh, M., Hanna, G.B., Elson, D.S.: Intraoperative measurement of bowel oxygen saturation using a multispectral imaging laparoscope. Biomedical Optics Express 6, 4179-4190 (2015)
5. Clancy, N.T., Arya, S., Corbett, R., Singh, M., Stoyanov, D., Crane, J.S., Duncan, N., Ebner, M., Caro, C.G., Hanna, G., Elson, D.S.: Optical Measurement of Anastomotic Oxygenation Dynamics. In: Biomedical Optics 2014, BS3A.23. Optical Society of America, (2014)
6. Akbari, H., Halig, L.V., Schuster, D.M., Osunkoya, A., Master, V., Nieh, P.T., Chen, G.Z., Fei, B.: Hyperspectral imaging and quantitative analysis for prostate cancer detection. Journal of Biomedical Optics 17, 0760051-07600510 (2012)
7. Clancy, N.T., Lin, J., Arya, S., Hanna, G.B., Elson, D.S.: Dual multispectral and 3D structured light laparoscope. Multimodal Biomedical Imaging X, Proc. of SPIE Vol. 9316, 93160C (2015)
8. Clancy, N.T., Stoyanov, D., James, D.R.C., Di Marco, A., Sauvage, V., Clark, J., Yang, G.-Z., Elson, D.S.: Multispectral image alignment using a three channel endoscope *in vivo* during minimally invasive surgery. Biomedical Optics Express 3, 2567-2578 (2012)
9. Long, J., Shelhamer, E., Darrell, T.: Fully Convolutional Networks for Semantic Segmentation. The IEEE Conference on Computer Vision and Pattern Recognition (CVPR) (2015)
10. Ronneberger, O., Fischer, P., Brox, T.: U-Net: Convolutional Networks for Biomedical Image Segmentation. 18th International Conference on Medical Image Computing and Computer Assisted Interventions (MICCAI) Part III, 234-241 (2015)
11. Clancy, N.T., Stoyanov, D., Maier-Hein, L., Groch, A., Yang, G.-Z., Elson, D.S.: Spectrally encoded fiber-based structured lighting probe for intraoperative 3D imaging. Biomedical Optics Express 2, 3119-3128 (2011)
12. Maier-Hein, L., Groch, A., Bartoli, A., Bodenstedt, S., Boissonnat, G., Chang, P.L., Clancy, N.T., Elson, D.S., Haase, S., Heim, E., Hornegger, J., Jannin, P., Kenngott, H., Kilgus, T., Muller-Stich, B., Oladokun, D., Rohl, S., dos Santos, T.R., Schlemmer, H.P., Seitel, A., Speidel, S., Wagner, M., Stoyanov, D.: Comparative Validation of Single-Shot Optical Techniques for Laparoscopic 3-D Surface Reconstruction. Medical Imaging, IEEE Transactions on 33, 1913-1930 (2014)
13. Lin, J., Clancy, N.T., Elson, D.S.: An endoscopic structured light system using multispectral detection. Int J CARS 10, 1941-1950 (2015)
14. Lin, J., Clancy, N.T., Stoyanov, D., Elson, D.S.: Tissue Surface Reconstruction Aided by Local Normal Information Using a Self-calibrated Endoscopic Structured Light System. 18th International Conference on Medical Image Computing and Computer Assisted Interventions (MICCAI) Part III, 405-412 (2015)
15. Jia, Y., Shelhamer, E., Donahue, J., Karayev, S., Long, J., Girshick, R., Guadarrama, S., Darrell, T.: Caffe: Convolutional Architecture for Fast Feature Embedding. Proceedings of the 22nd ACM international conference on Multimedia, 675-678. ACM, Orlando, Florida, USA (2014)